%% file: aaai25.tex
\documentclass[letterpaper]{article} 
\usepackage{aaai25}  
\usepackage{times}  
\usepackage{helvet}  
\usepackage{courier}  
\usepackage[hyphens]{url}  
\usepackage{graphicx} 
\usepackage{natbib}  
\usepackage{caption} 
\usepackage{algorithm}
\usepackage{algorithmic}
\usepackage{multirow}
\usepackage{booktabs}
\usepackage{amssymb}
\usepackage{amsmath} 
\usepackage{paralist}

\usepackage{newfloat}
\usepackage{listings}
\DeclareCaptionStyle{ruled}{labelfont=normalfont,labelsep=colon,strut=off} 
\floatstyle{ruled}
\newfloat{listing}{tb}{lst}{}
\floatname{listing}{Listing}
\title{APAR: Modeling Irregular Target Functions in Tabular Regression via Arithmetic-Aware Pre-Training and Adaptive-Regularized Fine-Tuning}
\author{
    Hong-Wei Wu, Wei-Yao Wang, Kuang-Da Wang, Wen-Chih Peng
}
\affiliations{
    National Yang Ming Chiao Tung University, Hsinchu, Taiwan \\
    johnnyhwu.cs11@nycu.edu.tw, sf1638.cs05@nctu.edu.tw, gdwang.cs10@nycu.edu.tw, wcpeng@cs.nycu.edu.tw
}

\usepackage{bibentry}

\begin{document}

\maketitle

\begin{abstract}
Tabular data are fundamental in common machine learning applications, ranging from finance to genomics and healthcare.
This paper focuses on tabular regression tasks, a field where deep learning (DL) methods are not consistently superior to machine learning (ML) models due to the challenges posed by irregular target functions inherent in tabular data, causing sensitive label changes with minor variations from features.
To address these issues, we propose a novel \textbf{A}rithmetic-Aware \textbf{P}re-training and \textbf{A}daptive-\textbf{R}egularized Fine-tuning framework (APAR), which enables the model to fit irregular target function in tabular data while reducing the negative impact of overfitting.
In the pre-training phase, APAR introduces an arithmetic-aware pretext objective to capture intricate sample-wise relationships from the perspective of continuous labels.
In the fine-tuning phase, a consistency-based adaptive regularization technique is proposed to self-learn appropriate data augmentation.
Extensive experiments across 10 datasets demonstrated that APAR outperforms existing GBDT-, supervised NN-, and pretrain-finetune NN-based methods in RMSE (+9.43\% $\sim$ 20.37\%), and empirically validated the effects of pre-training tasks, including the study of arithmetic operations. Our code and data
are publicly available at https://github.com/johnnyhwu/APAR.
\end{abstract}

%

\section{Introduction}

\input{src/intro}

\section{Related Work}
\input{src/related}

\section{Problem Formulation}
\input{src/problem}
\section{The Proposed Approach}
\input{src/solution/overview}
\input{src/solution/model_arch}
\input{src/solution/pretrain}
\input{src/solution/finetune}

\section{Experiments}
\input{src/exp/overview}

\input{src/exp/setup}
\input{src/exp/rq1}
\input{src/exp/rq2}
\input{src/exp/rq3}
\input{src/exp/rq4}

\section{Conclusion and Future Works}
\input{src/conclusion}
\bibliography{aaai25}

\newpage
\clearpage
\appendix
\input{src/appendix}
\end{document}

%% file: src/intro.tex
\begin{figure*}[t]
    \centering
    \includegraphics[width=0.82\textwidth]{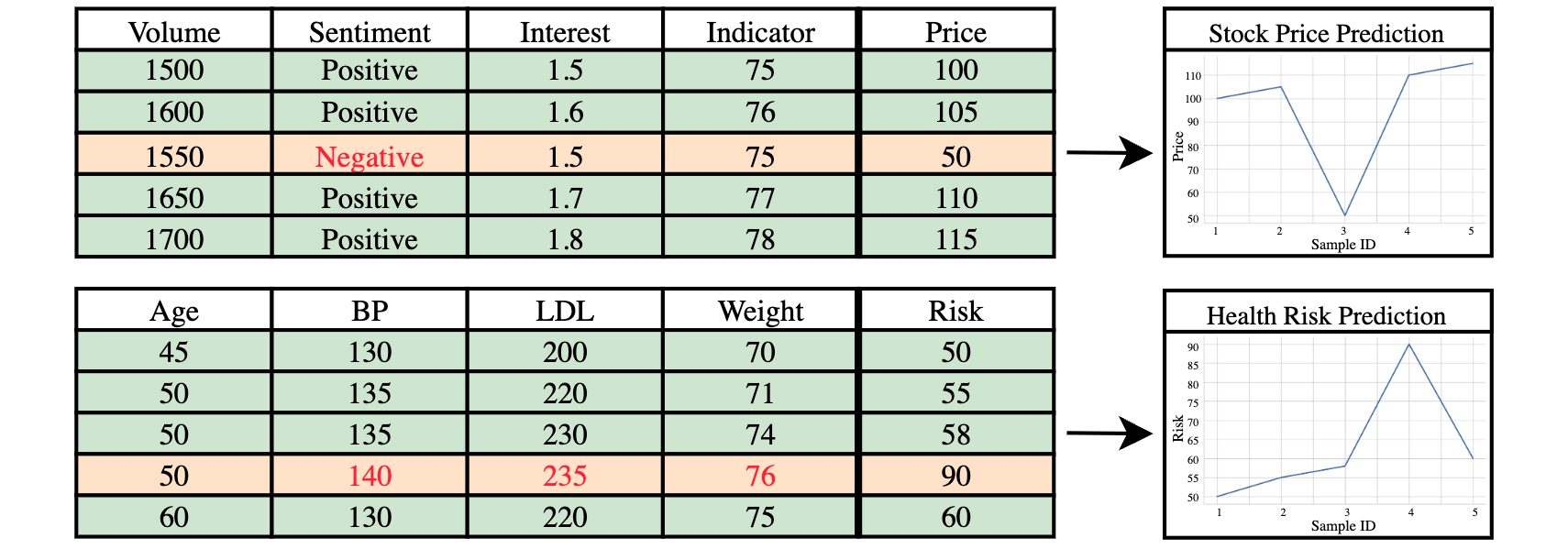}
    \caption{
    Illustrations of the impacts of irregular target functions commonly found in tabular regression tasks for finance (stock price prediction) and medical (health risk prediction) data. Small changes in features (marked in red) can lead to significant changes in the target variable.}
    \label{fig:irregular_target_function}
\end{figure*}



The tabular regression task, prevalent in sectors such as healthcare \cite{rao2023machine,jain2024predicting} and finance \cite{DBLP:conf/cikm/DuWP23,deng2024forecasting}, has commonly been addressed using Gradient Boosting Decision Tree (GBDT) models (e.g., CatBoost \cite{prokhorenkova2018catboost}).
Despite recent advancements in neural networks (NNs), they often fail to consistently outperform GBDT models in this domain \cite{DBLP:journals/corr/abs-2402-01204}.
This is attributed to the understanding of important features from distinct characteristics of tabular data, such as feature heterogeneity and the presence of uninformative features, which make it challenging to identify important features.
On the other hand, the irregular target functions prevent NNs from learning high-frequency components of heterogeneous tabular datasets, and negatively impact NN performance due to overfitting \cite{DBLP:conf/nips/BeyazitKLWF23}.

Prior research \cite{gorishniy2021revisiting,yan2023t2g,chen2023excelformer} has primarily focused on addressing feature heterogeneity and uninformative features; however, the issue of irregular target functions remains relatively unexplored, especially in the context of tabular regression tasks having continuous labels instead of explicit boundaries between labels.
Irregular target functions play a critical role since minor deviations in input features lead to major changes in target values \cite{DBLP:conf/nips/BeyazitKLWF23}.
For instance, as illustrated in Figure \ref{fig:irregular_target_function}, in a medical scenario (e.g., health risk prediction), a slight variation in a patient's weight crossing a specific threshold can significantly alter the corresponding health status.
In addition, the stock price is degraded significantly due to only the sentiment change (e.g., by wars).
This phenomenon is less common in other data modalities; for example, a minor change in a single pixel in an image is unlikely to change its appearance. 
As NNs need to accurately model irregular target functions in tabular regression tasks while suffering from overfitting, it is crucial to emphasize the significance of advancing tabular regression methods capable of modeling sensitive changes between tabular features and labels.

Therefore, we focus on learning to fit irregular target functions for \textit{tabular regression} tasks.
Prior studies have mitigated this problem by addressing it from two perspectives: preventing overfitting on samples \cite{kossen2021self,ucar2021subtab, wang2022transtab} and features \cite{yoon2020vime,arik2021tabnet,somepalli2021saint}.
However, they are inferior in terms of utilizing label information due to the significant sparsity for regression labels (e.g., supervised contrastive learning for tabular classification tasks \cite{DBLP:journals/corr/abs-2404-17489}) as well as in corrupting important features that are related to predictions (e.g., random feature masks \cite{DBLP:journals/corr/abs-2310-18541}).



To address the aforementioned challenges, we propose a novel \textbf{A}rithmetic-aware \textbf{P}re-training and \textbf{A}daptive-\textbf{R}egularized Fine-tuning framework (APAR) for tabular regression tasks, consisting of the pretrain-finetune strategy for modeling irregular target functions.
Specifically, a Transformer-based \cite{vaswani2017attention} backbone is adopted with a tabular feature tokenizer to encode tabular heterogeneity.
In the pre-training phase, an arithmetic-aware task is introduced to learn sample-wise relationships by predicting the combined answer of arithmetic operations on continuous labels. 
In the fine-tuning phase, we propose an adaptive regularization technique to reinforce the model to self-learn proper data augmentation based on feature importance by training the model to understand similar representations between original and augmented data.
We compared our APAR with GBDT and supervised as well as pretrain-finetune NNs on 10 datasets, which demonstrated a significant improvement of at least 9.43\% in terms of the RMSE score compared with the state-of-the-art baseline.

In brief, our main contributions are described as follows:
\begin{compactitem}
\item We present a principle recipe that enables NNs to effectively perform tabular regression tasks by addressing irregular target functions with the advantage of continuous labels and mitigating the negative impacts of overfitting on both samples and features.
\item We introduce an arithmetic-aware pre-training method to learn the relationships between samples by solving arithmetic operations from continuous labels. 
In addition, our adaptive-regularized fine-tuning technique allows the model to perform self-guided data augmentation, offering effective regularization and generalization in downstream tasks.
\item Extensive experiments across 10 datasets were systematically conducted to demonstrate an improvement from 9.43\% to 20.37\% compared with the GBDT-, supervised NN-, and pretrain-finetune NN-based methods.
\end{compactitem}

%% file: src/related.tex
\begin{figure*}[h]
    \centering
    \includegraphics[width=0.9\textwidth]{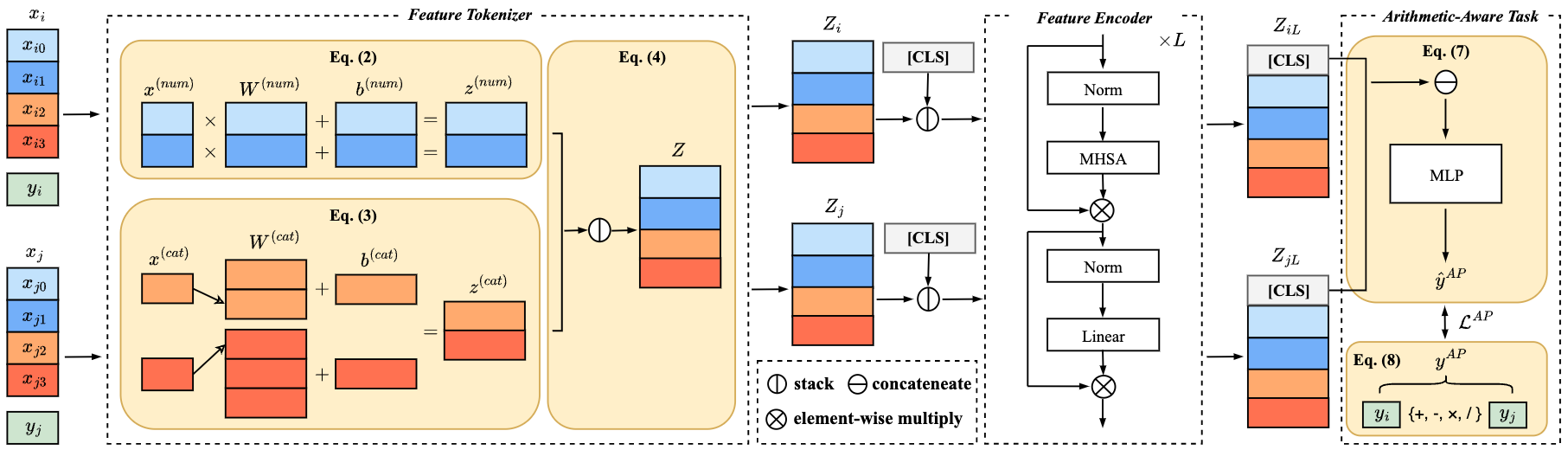}
    \caption{
    Illustration of the \textit{Arithmetic-Aware Pre-Train} phase of APAR. Sample pairs are processed through the \textit{Feature Tokenizer} and \textit{Feature Encoder}, the outputs of which are concatenated for arithmetic prediction, enabling the model to understand inter-sample relationships in tabular regression.}
    \label{fig:pretrain}
\end{figure*}

Recently, deep learning approaches for tabular data have demonstrated effective performance.
For instance, \citet{song2019autoint} employed multi-head self-attention for intra-sample feature interactions. 
\citet{huang2020tabtransformer} and \citet{gorishniy2021revisiting} adapted the Transformer architecture to tabular data, targeting both categorical and numerical features. 
Nonetheless, GBDT-based approaches \cite{chen2016xgboost,ke2017lightgbm,prokhorenkova2018catboost} are still competitive in tabular benchmarks due to the highly complex and heterogeneous characteristics of tabular data, causing NNs to overfit on irregular target functions.

To prevent NN from overfitting on samples, prior research has proposed to considering sample-wise relationships while learning individual representations. 
NPT \cite{kossen2021self} and SAINT \cite{somepalli2021saint} utilize self-attention to explicitly reason about relationships between samples.
Similarly, \citet{ucar2021subtab} employed self-supervised contrastive loss to ensure that the model outputs similar representations for different feature subsets of the same sample, and dissimilar representations for subsets of different samples. 
Although these approaches effectively capture relationships between samples, they require relatively large batch sizes (e.g., 4096 in NPT) to include more samples computed under self-attention, leading to high computational and memory consumption. 
Moreover, they do not leverage supervised labels to learn contextualized representations based on explicit information.
To incorporate supervised lavels, Supervised Contrastive Loss \cite{khosla2020supervised} and TransTab \cite{wang2022transtab} enable models to learn improved sample representations by making samples with the same label similar, and samples with different labels dissimilar.
Similarly, \citet{DBLP:journals/corr/abs-2404-17489} improved positive samples by augmenting anchors with features from samples of the same class to learn better representations based on explicit labels. 
Nonetheless, these approaches rely on discrete labels to determine positive or negative samples, which becomes extremely sparse for regression tasks where continuous labels are used.

To prevent NN from overfitting features, another line of research has utilized regularization techniques to avoid the model paying too much attention to a single feature.
For instance, VIME \cite{yoon2020vime}, SAINT \cite{somepalli2021saint}, TabNet \cite{arik2021tabnet} and ReConTab \cite{DBLP:journals/corr/abs-2310-18541} use feature corruption, which encourages the model to output consistent predictions when features are randomly masked or mixed-up with other samples.
However, randomly masking or mixing-up might inadvertently corrupt important features, causing the model to learn to predict based on uninformative features, and thus deteriorating learning representations.
On the other hand, our proposed approach incorporates arithmetic from continuous labels as the pre-training task, and adaptively self-learns proper regularization during the fine-tuning stage.

%% file: src/problem.tex
In this paper, we focus on regression tasks within the tabular domain.
A dataset is denoted as $D=\{(x_i, y_i)\}_{i=1}^n$, where $x_i=(x_{i1}^{T}, ..., x_{ik}^{T}) \in \mathbb{R}^{k}$ represents an object consisting of $k$ features, with $T \in \{num, cat\}$ indicating whether the feature is numerical or categorical.
The corresponding label is denoted as $\displaystyle y_i \in \mathbb{R}$.
Given an input sample $x$, our goal is to learn a representation $Z$ of the sample that effectively encapsulates both feature-wise and sample-wise information that is able to maintain robustness against heterogeneous and uninformative features to precisely predict the corresponding target $y$.

%% file: src/solution/overview.tex

The \textbf{APAR} framework employs a pretrain-finetune framework, outlined in Figures \ref{fig:pretrain} (pre-training phase) and \ref{fig:finetune} (fine-tuning phase).
Our APAR framework consists of two modules: the \textit{Feature Tokenizer} and \textit{Feature Encoder}.
In the pre-training stage, a pair of two samples are encoded by the feature tokenizer and feature encoder to obtain their corresponding representations, and the contextualized [CLS] token is used for predicting arithmetic outcomes based on their numerical labels.
In the fine-tuning stage, a test sample is augmented by applying it with a gate vector.
Both the original and augmented samples are then encoded by the pre-trained feature tokenizer and feature encoder, generating two contextualized [CLS] tokens.
The model is trained to adapt the gate vector based on self-learned feature importance, ensuring consistent predictions across the original and augmented samples.

%% file: src/solution/model_arch.tex
\subsection{Model Architecture}




\subsubsection{Feature Tokenizer}


To transform input features into representations, the feature tokenizer is introduced to convert categorical and numerical features of a sample into a sequence of embeddings.
Similar to \cite{grinsztajn2022tree}, the feature tokenizer can prevent the rotational invariance of NNs by learning distinct embeddings for each feature.

Given $j$-th feature $x_{ij}$ of the $i$-th sample, the tokenizer generates a $d$-dimensional embedding $z_{ij} \in \mathbb{R}^{d}$. 
Formally, the embedding is computed as:
\begin{equation}
z_{ij} = b + f(x_{ij}) \in \mathbb{R}^{d},
\end{equation}
where $b$ is a bias term and $f$ represents a transformation function.
For numerical features, $f$ involves an element-wise product with a weighting vector $W^{(num)} \in \mathbb{R}^d$:
\begin{equation}
z_{ij}^{(num)} = b_{j}^{(num)} + x_{ij}^{(num)} \cdot W_{j}^{(num)} \in \mathbb{R}^{d}.
\end{equation}

For categorical features, $f$ applies a lookup in the embedding matrix $W^{(cat)} \in \mathbb{R}^{c \times d}$, where $c$ is the number of categories, and $e_{ij}$ is a one-hot vector for the corresponding categorical feature: 
\begin{equation}
z_{ij}^{(cat)} = b_{j}^{(cat)} + e_{ij} \cdot W_{j}^{(cat)} \in \mathbb{R}^{d}.
\end{equation}

Finally, the output from the feature tokenizer $Z_i$ is concatenated by the embeddings of all features of a sample $x_i$:
\begin{equation}
Z_{i} = \text{stack}[z_{i1}, ..., z_{ik}] \in \mathbb{R}^{k \times d}.
\end{equation}

\subsubsection{Feature Encoder}
Since our aim is to explore the strategies of the pre-training and fine-tuning stages similar to \cite{huang2020tabtransformer,somepalli2021saint}, the Transformer blocks encompassing multi-head self-attention and feed-forward networks are adopted as the feature encoder to encode intricate interrelations among heterogeneous and unformative features and to align the comparison.

Specifically, the embedding of a [CLS] token is first appended to the output $Z_i$ of the feature tokenizer, which is then fed into $L$ Transformer layers, $F_1, ..., F_L$:
\begin{equation}
\begin{aligned}
&Z_{i0} = \text{stack}[\text{[CLS]}, Z_i], \\
&Z_{il} = F_{l}(Z_{i(l-1)}); l = 1, 2, ..., L.
\end{aligned}
\end{equation}
The output of the encoder can then be used to learn contextualized knowledge from the pre-training stage and downstream tasks from the fine-tuning stage.






\begin{figure*}[h]
    \centering
    \includegraphics[width=0.9\textwidth]{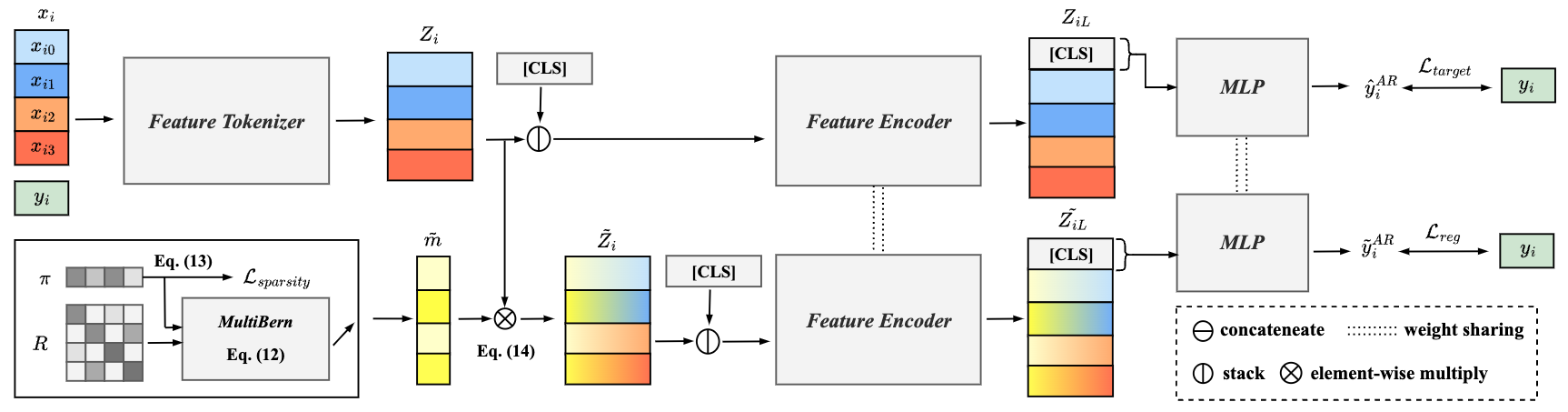}
    \caption{Illustration of the \textit{Adaptive Regularization Fine-Tuning} phase of APAR. In this phase, an input sample is processed through the \textit{Feature Tokenizer} to generate feature embeddings, which are augmented using a dynamically adaptive gate vector. The model is trained to predict consistent labels from varying inputs, which enhances the model's robustness to uninformative features and performance on the target task.}
    
    \label{fig:finetune}
\end{figure*}

%% file: src/solution/pretrain.tex
\subsection{Arithmetic-Aware Pre-Training}
\label{sec:aridaformer_pretrain}

The goal of the pre-training phase is to integrate sample-wise information into the representation of each sample; however, existing methods such as supervised contrastive learning \cite{khosla2020supervised,wang2022transtab,DBLP:journals/corr/abs-2404-17489} are ineffective in regression scenarios due to their reliance on discrete class labels, as opposed to regression's continuous labels.
Also, simply relying on attention mechanisms (e.g., \cite{kossen2021self,somepalli2021saint}) underutilizes the relationship between label information across samples.
To that end, we introduce a novel arithmetic-aware pretext task by conditioning continuous labels. 

Analogous to solving for an unknown in a set of simultaneous equations in mathematics, our pre-training goal is to introduce constraints that are able to narrow the possible outcomes (i.e., search space) of the unknown.
Therefore, the arithmetic-aware pre-training task is proposed to enable the model to discern relationships between samples by utilizing continuous labels in tabular regression. 
Intuitively, pairing sample A with different samples B, C, and D generates unique aggregated outcomes, such as A+B, A+C, and A+D. 
These pairs impose constraints and guide the model in learning a fine-grained representation of A that considers the context from not only itself but also other paired samples.
In our work, we opt for a simple yet effective pre-training task by incorporating an arithmetic operator with two samples at a time as the pretext objective.

As shown in Figure \ref{fig:pretrain}, the arithmetic-aware pre-training process starts by selecting two random samples, $x_i$ and $x_j$, from the dataset, each with corresponding labels $y_i$ and $y_j$. 
These samples undergo processing through the \textit{Feature Tokenizer} and \textit{Feature Encoder} to produce their respective representations, $Z_{iL}$ and $Z_{jL}$:
\begin{equation} \label{eq:6}
\begin{aligned}
Z_{i|j} &= \textit{FeatureTokenizer}(x_{i|j}), \\
Z_{iL|jL} &= \textit{FeatureEncoder}(\text{stack}[[\text{CLS}], Z_{i|j}]), \\
\end{aligned}
\end{equation}
where $i|j$ indicates the term is either the $i$- or $j$-th sample.
Subsequently, the representations of the [CLS] token $Z_{iL}^{\text{[CLS]}}$ and $Z_{jL}^{\text{[CLS]}}$ are extracted from $Z_{iL}$ and $Z_{jL}$, respectively. They are then concatenated and fed into a Multilayer Perceptron (MLP) to predict the outcome $\hat{y}^{\text{AP}}$ of the arithmetic operation:
\begin{equation}
\begin{aligned}
\hat{y}^{\text{AP}} &= \text{MLP}(\text{concat}[Z_{iL}^{[CLS]}, Z_{jL}^{[CLS]}]).
\end{aligned}
\end{equation}

The pre-training task involves applying arithmetic operations on the sample labels $y_i$ and $y_j$, including addition, subtraction, multiplication, and division\footnote{We empirically chose arithmetic operations in our experiments (See Sec. 5.5 for detailed analyses).}. The resulting ground truth $y^{\text{AP}}$ for the arithmetic task is represented as: 
\begin{equation}
\begin{aligned}
y^{\text{AP}} =
\begin{cases}
y_i + y_j, & \text{for addition}; \\
y_i - y_j, & \text{for subtraction}; \\
y_i \times y_j, & \text{for multiplication}; \\
y_i / y_j, & \text{for division}.
\end{cases}
\end{aligned}
\end{equation}
Finally, the model is then trained to minimize $\mathcal{L}^{\text{AP}}$:
\begin{equation}
\begin{aligned}
\mathcal{L}^{\text{AP}} = \frac{1}{n} \sum^{n}(y^{\text{AP}} - \hat{y}^{\text{AP}})^2.
\end{aligned}
\end{equation}

This pre-training task embeds an awareness of arithmetic relationships between samples into the model, thereby enabling it to adeptly handle the irregularities of target functions. The detailed procedure of arithmetic-aware pre-training is summarized in Algorithm \ref{alg:pre-train} in Appendix \ref{appendix:pretrain_algo}.

%% file: src/solution/finetune.tex
\subsection{Adaptive-Regularized Fine-Tuning}
\label{sec:aridaformer_finetune}

In the fine-tuning phase of APAR, we introduce an adaptive-regularized fine-tuning method that adaptively augments samples by considering feature importance and their correlated structure.
As shown in Figure \ref{fig:finetune}, an input sample is processed by the pre-trained feature tokenizer to produce feature embeddings, which are subsequently augmented using a dynamically adaptive gate vector.
The model is fine-tuned to predict a consistent label from these variant inputs, which enables the model to perform data augmentation, guided by the model-learned importance of each feature, to improve performance on the downstream task.

\subsubsection{Adaptive Learning}

When learning the importance of each feature, we consider the correlation structure of the features rather than assuming independence, as the feature selection process is influenced by these correlations \cite{katrutsa2017comprehensive}.
Specifically, a correlated gate vector for augmenting the input sample is generated from a multivariate Bernoulli distribution where the mean is determined by learnable parameters reflecting feature importance.
The distribution is jointly updated when fine-tuning the model to adaptively utilize the self-learned gate vector to augment the input sample.

Specifically, given the correlation matrix $R \in [-1, 1]^{k \times k}$ representing the correlation structure, the Gaussian copula is defined as:
\begin{equation}
C_R(U_1, ..., U_k) = \Phi_R(\Phi^{-1}(U_1), ..., \Phi^{-1}(U_k)),
\end{equation}
where $\Phi_R$ denotes the joint cumulative distribution function (CDF) of a multivariate Gaussian distribution with a mean zero vector and correlation matrix $R$, $\Phi^{-1}$ is the inverse CDF of the standard univariate Gaussian distribution, and $U_j \sim Uniform(0, 1)$ for $j \in [k]$.

Afterwards, we sample a gate vector $m$ to augment the input sample from a multivariate Bernoulli distribution that maintains the correlation structure of the input features as:
\begin{equation}
m \sim MultiBern(\pi; R).
\end{equation}
Formally, $m_j$ is set to 1 if $U_j \le \pi_{j}$ and 0 if $U_j > \pi_{j}$, for $j \in [k]$. Here, $\pi$ represents a set of learnable parameters indicating feature importance.
For differentiability, we apply the reparametrization trick \cite{wang2020relaxed}, resulting in a relaxed gate vector $\tilde{m}$ from the following equation:
\begin{equation}
\tilde{m}_j = \sigma(\frac{1}{\tau}(\log \pi_{j} - \log (1-\pi_{j}) + \log U_{j} - \log (1-U_{j}))),
\end{equation}
where $\sigma(x) = (1 + \exp(-x))^{-1}$ is the sigmoid function, and a temperature parameter $\tau \in (0, \infty)$.

The relaxed gate vector $\tilde{m}$ is then used to be multiplied with the feature embeddings for data augmentation. The feature selection probability $\pi$ is learned during training and is adjusted according to the performance of the model. To induce the sparsity of the selected features, the sparsity loss, $\mathcal{L}_{\text{sparsity}}$ is calculated as follows:
\begin{equation}
\mathcal{L}_{\text{sparsity}} = \sum_{i}^{k} \pi_{i}.
\end{equation}

\subsubsection{Fine-Tuning Loss Function}
As depicted in Figure \ref{fig:finetune}, for each input sample $x_{i}$ with the label $y_{i}$, it is processed by the feature tokenizer and feature encoder to obtain the representation of the sample $Z_{iL}$, as detailed in Equation (\ref{eq:6}).
Simultaneously, the original feature embeddings $Z_{i}$ are element-wise multiplied with the relaxed gate vector $\tilde{m}$ to obtain augmented feature embeddings $\tilde{Z}_{i}$, as shown below:

\begin{equation}
\tilde{Z}_{i} = Z_{i} \odot \tilde{m}.
\end{equation}

These augmented embeddings $\tilde{Z}_{i}$ are then stacked with the [CLS] token and input into the feature encoder to obtain the augmented representation $\tilde{Z}_{iL}$:

\begin{equation}
\tilde{Z}_{iL} = \textit{FeatureEncoder}(\text{stack}[[\text{CLS}], \tilde{Z}_{i}]).
\end{equation}

The representation of the [CLS] token $Z_{iL}^{[\text{CLS}]}$ and $\tilde{Z}_{iL}^{[\text{CLS}]}$, extracted from $Z_{iL}$ and $\tilde{Z}_{iL}$, respectively, are input into an MLP to generate the corresponding predictions $\hat{y}_{i}^{\text{AR}}$ and $\tilde{y}_{i}^{\text{AR}}$, as described by the following equations:

\begin{equation}
\hat{y}_{i}^{\text{AR}} = \text{MLP}(Z_{iL}^{[\text{CLS}]}), \tilde{y}_{i}^{\text{AR}} = \text{MLP}(\tilde{Z}_{iL}^{[\text{CLS}]}).
\end{equation}

The losses for the target task $\mathcal{L}_{\text{target}}$ and the regularization $\mathcal{L}_{\text{reg}}$ are computed as follows:

\begin{equation}
\mathcal{L}_{\text{target}} = \frac{1}{n} \sum^{n}_{i}(y_i - \hat{y}_{i}^{\text{AR}})^2, \mathcal{L}_{\text{reg}} = \frac{1}{n} \sum^{n}_{i}(y_i - \tilde{y}_{i}^{\text{AR}})^2.
\end{equation}

During the fine-tuning phase, the total loss, $\mathcal{L}^{\text{AR}}$, is the weighted sum of the target task loss, regularization loss, and feature sparsity loss, defined as:
\begin{equation}
\mathcal{L}^{\text{AR}} = \alpha \mathcal{L}_{\text{target}} + \beta \mathcal{L}_{\text{reg}} + \gamma \mathcal{L}_{\text{sparsity}},
\end{equation}
where $\alpha$, $\beta$, and $\gamma$ are hyperparameters within the range $[0, 1]$. The procedure of adaptive-regularized fine-tuning is summarized in Algorithm \ref{alg:fine-tune} in Appendix \ref{appendix:fintune_algo}.

%% file: src/exp/overview.tex
\begin{table}
  \begin{center}
    \caption{Statistics of each dataset.}
    \label{tab:dataset_overview}
    \small

        \begin{tabular}{lccccc}
          \toprule
          & \textbf{BD} & \textbf{AM} & \textbf{HS} & \textbf{GS} & \textbf{ER} \\
          \midrule
          \text{\#instances} & \text{73203} & \text{60786} & \text{61784} & \text{36733} & \text{21643} \\
          \text{\#num feats} & \text{230} & \text{230} & \text{230} & \text{10} & \text{23} \\
          \text{\#cat feats} & \text{17} & \text{17} & \text{17} & \text{0} & \text{2} \\
          \midrule
          \midrule
          & \textbf{PM} & \textbf{BS} & \textbf{YE} & \textbf{KP} & \textbf{FP} \\
          \midrule
          \text{\#instances} & \text{41757} & \text{17389} & \text{515345} & \text{241600} & \text{300153} \\
          \text{\#num feats} & \text{11} & \text{7} & \text{90} & \text{0} & \text{2} \\
          \text{\#cat feats} & \text{1} & \text{8} & \text{0} & \text{14} & \text{7} \\
          \bottomrule
        \end{tabular}
  \end{center}
\end{table}

\begin{table*}
\small
  \begin{center}
    \caption{Quantitative results of all groups of baselines and our proposed APAR. For each dataset, the best result in each column is in boldface, while the second best result is underlined. * denotes without pre-training.}
    \label{tab:exp1_result}
    \begin{tabular}{cccccccccccc|c}
    
      \toprule
      
      \text{Group} & & \text{BD} & \text{AM} & \text{HS} & \text{GS} & \text{ER} & \text{PM} & \text{BS} & \text{YE} & \text{KP} & \text{FP} & \text{Rank}\\

      \midrule

      \multirow{3}{*}{\text{GBDT-based}} & \text{XGB} & \text{0.2476} & \text{0.2472} & \text{0.3509} & \text{0.1489} & \text{0.0510} & \text{{0.6075}} & \underline{\text{0.0244}} & \text{0.2166} & \text{0.2559} & \text{0.4066} & \text{6.0} \\
      & \text{LGBM} & \text{0.2506} & \text{0.2429} & \text{0.3354} & \text{0.1575} & \text{0.0693} & \text{0.7445} & \text{0.0458} & \text{0.2156} & \text{0.3718} & \text{0.4213} & \text{6.5} \\
      & \text{CB} & \text{0.2406} & \text{0.2441} & \text{0.3423} & \text{0.1526} & \text{0.0557} & \text{0.7398} & \text{0.0469} & \text{0.2175} & \text{0.3087} & \text{0.4460} & \text{6.3} \\
      
      \midrule
      
      \multirow{4}{*}{\shortstack{Supervised\\NN-based}} & \text{MLP} & \text{0.2728} & \text{0.2617} & \text{0.4314} & \text{0.1743} & \text{0.0499} & \text{0.6973} & \underline{\text{0.0244}} & \text{0.2179} & \textbf{0.0500} & \text{0.3659} & \text{6.6} \\
       & \text{AutoInt} & \text{0.2498} & \text{0.2456} & \text{0.3510} & \text{0.1755} & \text{0.1549} & \text{0.7790} & \text{0.0974} & \text{0.2161} & \text{0.0830} & \text{0.3843} & \text{7.7} \\
       & \text{FT-T} & \text{0.2452} & \underline{\text{0.2352}} & \text{0.3457} & \text{0.1710} & \text{0.0538} & \text{0.8160} & \text{0.0591} & \text{0.2163} & \underline{\text{0.0547}} & \underline{\text{0.3421}} & \text{5.5} \\
       & \text{TabNet*} & \text{0.2435} & \text{0.2502} & \text{0.3467} &  \underline{\text{0.1249}} & \text{0.0838} & \underline{\text{0.5537}} & \text{0.0591} & \textbf{0.2114} & \text{0.1849} & \text{0.3657} & \text{5.1} \\

       \midrule

       \multirow{2}{*}{\shortstack{NN-based with a\\Pretrain-Finetune}} & \text{VIME} & \text{0.2412} & \text{0.2606} & \text{0.3746} & \text{0.1266} & \underline{\text{0.0422}} & \text{0.6557} & \text{0.1360} & \text{0.2184} & \text{0.1766} & \text{0.3685} & \text{6.4}\\
       & \text{TabNet} & \underline{\text{0.2404}} & \text{0.2427} & \underline{\text{0.3333}} & \text{0.1352} & \text{0.0846} & \text{0.5880} & \text{0.0479} & \underline{\text{0.2126}} & \text{0.0566} & \text{0.3562} & \underline{\text{3.8}}
       \\
       
       \midrule
       
       & \text{APAR} & \textbf{0.2397} & \textbf{0.2293} & \textbf{0.3305} & \textbf{0.1205} & \textbf{0.0338} & \textbf{0.5239} & \textbf{0.0139} & \text{0.2148} & \textbf{0.0500} & \textbf{0.3303} & \text{\textbf{1.3}}\\
       
      \bottomrule
    \end{tabular}
  \end{center}
\end{table*}

In this section, we attempt to answer the following research questions on a wide range of real-world datasets:
\begin{compactitem}
    \item \textbf{RQ1:} Does our proposed framework, APAR, outperform the existing NN-based and GBDT-based approaches?
    \item \textbf{RQ2:} How does the performance of the proposed arithmetic-aware pre-training task compare to other pre-training approaches in tabular regression?
    \item \textbf{RQ3:} Does the adaptive regularization enhance the model's performance during the fine-tuning phase?
    \item \textbf{RQ4:} How do different arithmetic operations affect the performance across various scenarios?
\end{compactitem}

%% file: src/exp/setup.tex
\subsection{Experimental Setup}


\noindent\textbf{Datasets Overview.}
In our experiments, we utilized 10 publicly real-world datasets across diverse tabular regression applications (i.e., property valuation, environmental monitoring, urban applications, and performance analysis), spanning a range of scales from large-scale (over 100K samples) to medium-scale (50K to 100K samples) and small-scale (less than 50K samples). 
The datasets include Taiwan Housing (BD, AM, HS) \cite{tw_real_estate_platform} consisting of three building types (building, apartment, and house), Gas Emission (GS) \cite{misc_gas_turbine_co_and_nox_emission_data_set_551}, Election Results (ER) \cite{misc_real-time_election_results:_portugal_2019_513}, Beijing PM2.5 (PM) \cite{misc_beijing_pm2.5_381}, Bike Sharing (BS) \cite{misc_bike_sharing_dataset_275}, Year (YE) \cite{misc_yearpredictionmsd_203}, Kernel Performance (KP) \cite{misc_sgemm_gpu_kernel_performance_440}, and Flight Price (FP) \cite{flight_price_kaggle}.
Each dataset presents unique characteristics, including variations in features and label ranges.
A summary of the dataset characteristics is presented in Table \ref{tab:dataset_overview}, and we follow their corresponding protocols to split training, validation, and test sets.

Categorical features of each dataset were processed through label encoding \cite{hancock2020survey}, except in CatBoost \cite{prokhorenkova2018catboost} where built-in categorical feature support was used, while continuous features and labels were transformed using logarithmic scaling \cite{changyong2014log}.
For NN-based approaches, we utilize uniformly dimensioned embeddings for all categorical features.

\noindent\textbf{Baseline Methods.}
We compared APAR against several baselines categorized into three groups:
1) GBDT-based: \textbf{XGBoost (XGB)} \cite{chen2016xgboost}, \textbf{LightGBM (LGBM)} \cite{ke2017lightgbm} and \textbf{CatBoost (CB)} \cite{prokhorenkova2018catboost}.
2) Supervised NN-based: \textbf{MLP}, \textbf{AutoInt} \cite{song2019autoint}, and \textbf{FT-Transformer (FT-T)} \cite{gorishniy2021revisiting}.
3) NN-based with a pretrain-finetune strategy: \textbf{VIME} \cite{yoon2020vime} and \textbf{TabNet} \cite{arik2021tabnet}.

\noindent\textbf{Implementation Details.}
\label{sec:impl_details}
Our proposed APAR framework was developed using PyTorch version 1.13.1. The training was performed on an NVIDIA GeForce RTX 3090 GPU.
Regarding optimizers, we followed the original TabNet implementation by using the Adam optimizer \cite{kingma2014adam}. For all other models, we employed the AdamW optimizer \cite{loshchilov2018fixing} with $\beta_1 = 0.9$, $\beta_2 = 0.999$, and a weight decay of 0.01. A consistent StepLR scheduler was used for all deep learning models, and the batch size was set at 256 for each dataset and algorithm. Training continued until there was no improvement on the validation set for 10 consecutive epochs.

\noindent\textbf{Evaluation Metrics.}
Following \cite{gorishniy2021revisiting}, we used the root mean squared error (RMSE) for evaluating regression models. The ranks for each dataset were determined by sorting the scores obtained. 
The $Rank$ reflects the average rank across all datasets. 
All the results are the average of 5 different random seeds. 

\begin{table*}
\small
  \begin{center}
    \caption{Ablative experiments of different pre-training tasks (RQ2) and adaptive regularization (RQ3).}
    \label{tab:exp2_pretrain}
    \begin{tabular}{llcccccccccc}
      \toprule
      & & \text{BD} & \text{AM} & \text{HS} & \text{GS} & \text{ER} & \text{PM} & \text{BS} & \text{YE} & \text{KP} & \text{FP} \\
      \midrule
      \multirow{4}{*}{\text{RQ2}} & \text{w/o AP} & \text{0.2520} & \text{0.2377} & \text{0.3474} & \text{0.1549} & \text{0.0462} & \text{0.6173} & \text{0.0224} & \text{0.2175} & \text{0.0574} & \text{0.3603} \\
      & \text{AP $\rightarrow$ FR} & \text{0.2468} & \text{0.2512} & \text{0.3530} & \text{0.1259} & \text{0.0297} & \text{0.5758} & \text{0.0173} & \text{0.2148} & \text{0.0548} & \text{0.3409} \\
      & \text{AP $\rightarrow$ MR} & \text{0.2464} & \text{0.2464} & \text{0.3582} & \text{0.1956} & \text{0.1582} & \text{0.6403} & \text{0.0141} & \text{0.2198} & \text{0.0728} & \text{0.3718} \\
      & \text{AP $\rightarrow$ FR + MR} & \text{0.2508} & \text{0.2319} & \text{0.3530} & \text{0.1360} & \textbf{0.0266} & \text{0.5729} & \text{0.0140} & \textbf{0.2112} & \text{0.0548} & \text{0.3406} \\
      \midrule
      \text{RQ3} & \text{w/o AR} & \text{0.2536} & \text{0.2383} & \text{0.3501} & \text{0.1240} & \textbf{0.0266} & \text{0.5955} & \text{0.0632} & \text{0.2161} & \text{0.0520} & \text{0.3344} \\
      \midrule
      \midrule
      & \text{APAR (Ours)} & \textbf{0.2397} & \textbf{0.2293} & \textbf{0.3305} & \textbf{0.1205} & \textbf{0.0266} & \textbf{0.5239} & \textbf{0.0139} & \text{0.2148} & \textbf{0.0500} & \textbf{0.3303} \\
      \bottomrule
    \end{tabular}
  \end{center}
\end{table*}


\begin{table*}
\small
  \begin{center}
    \caption{Performance of using different arithmetic operations in Arithmetic-Aware Pre-Training. ``-'' indicates that the model did not converge during the pre-training phase.}
    \label{tab:exp3_arithmetic_operation}
    \begin{tabular}{lccccccccccc}
      \toprule
      & \text{BD} & \text{AM} & \text{HS} & \text{GS} & \text{ER} & \text{PM} & \text{BS} & \text{YE} & \text{KP} & \text{FP} & \text{Rank}\\
      \midrule
      \text{Addition} & \text{0.2495} & \textbf{0.2293} & \textbf{0.3305} & \text{0.1453} & \text{0.0338} & \textbf{0.5239} & \text{0.0182} & \text{0.2158} & \textbf{0.0500} & \textbf{0.3303} & \underline{\text{2.2}}\\
      \text{Subtraction} & \underline{\text{0.2408}} & \text{0.2413} & \text{0.3422} & \underline{\text{0.1250}} & \underline{\text{0.0287}} & \underline{\text{0.5783}} & \underline{\text{0.0147}} & \text{0.2199} & 
\underline{\text{0.0505}} & \text{0.3458} & \text{2.5}\\
      \text{Multiplication} & \textbf{0.2397} & \underline{\text{0.2317}} & \text{0.3458} & \textbf{0.1205} & \textbf{0.0266} & \text{0.5863} & \textbf{0.0139} & \textbf{0.2148} & \text{0.0506} & \underline{\text{0.3321}} & \textbf{1.9}\\
      \text{Division} & \text{0.2469} & \text{-} & \underline{\text{0.3375}} & \text{0.1420} & \text{-} & \text{-} & \text{-} & \underline{\text{0.2149}} & \text{0.0596} & \text{-} & \text{3.4}\\
      \bottomrule
    \end{tabular}
  \end{center}
\end{table*}

%% file: src/exp/rq1.tex
\subsection{Quantitative Results (RQ1)}
Table \ref{tab:exp1_result} presents the quantitative performance of APAR and the baselines.
Quantitatively, APAR was consistently superior to all approaches in overall ranking across 10 diverse datasets, achieving an average RMSE improvement of 9.18\% compared to the second-best ranking method.
We summarize the observations as follows:


\noindent\textbf{Selection of the Feature Encoder.}
We can observe that TabNet(*) and FT-Transformer demonstrate better performance compared with the other baselines in all three categories since they adopt Transformer architectures as their backbones to model intricate characteristics across tabular samples as well as features.
Nonetheless, the comparison of APAR, which employs the Transformer architecture in the feature encoder, and these baselines reveals the importance of considering the advantage of learning contextualized representations in a two-stage manner.


\noindent\textbf{Advantages of the Pretrain-Finetune Approach.} 
It is evident that comparing TabNet* with TabNet illustrates notable improvements with pre-training, demonstrating the value of the pretrain-finetune framework. 
However, VIME substantially hinders all performance due to not only the relatively simplified MLP architecture but also the lack of considering 
feature heterogeneity and rotational invariance, 
which again raises the need for leveraging the Transformer architecture with a feature tokenizer for tabular regression tasks. 
The effectiveness of our APAR highlights the capability of arithmetic-related pertaining tasks and adaptively learning contexts of features during the finetuning stage. 

%% file: src/exp/rq2.tex
\subsection{Effects of the Pre-Training Task (RQ2)}
To testify the design of the pre-training task in APAR, we evaluate arithmetic-aware pre-training with four variants: 1) remove (w/o AP), replacing it with 2) feature reconstruction (AP $\rightarrow$ FR), 3) mask reconstruction (AP $\rightarrow$ MR), and 4) AP $\rightarrow$ FR+MR.
Feature reconstruction is pre-trained to reconstruct with corrupt samples, which are randomly inserted constant values to some features. Mask reconstruction is pre-trained to predict the correct binary mask applied to the input sample, which aims to identify which parts of the input have been replaced.
As shown in Table \ref{tab:exp2_pretrain}, it is obvious that removing the pre-training task degrades the performance for all scenarios.
The deleterious effect of replacing our method with MR is due to the inadvertent masking of key features, which shifts the model's reliance to less relevant sample details and overlooks inter-sample relationships.
Although combining FR with MR improves performance, a significant gap remains compared to our arithmetic-aware pre-training task, indicating that utilizing the continuous labels in the regression scenario to design arithmetic tasks effectively encourages the model to consider inter-sample relationships.


%% file: src/exp/rq3.tex
\subsection{Effects of Adaptive Regularization (RQ3)}
To investigate the impact of incorporating adaptive-regularized fine-tuning in APAR, the performance of the removal of this design (w/o AR) was compared, as shown in the RQ3 row in Table \ref{tab:exp2_pretrain}.
Specifically, we fixed the target task loss weight $\alpha$ at 1 and optimized the regularization loss weight $\beta$ and sparsity loss weight $\gamma$ using the validation dataset. 
Removing the adaptive-regularized technique causes the model to be prone to overfitting on uninformative features, which degrades the performance across all datasets.
In contrast, APAR mitigates this limitation by adaptive regularization, leading to a substantial improvement.

%% file: src/exp/rq4.tex
\subsection{Variants of Arithmetic Operations (RQ4)}
We studied the performance of addition, subtraction, multiplication, and division across all datasets, as detailed in Table \ref{tab:exp3_arithmetic_operation}.
It can be seen that both addition and multiplication operations are more effective than subtraction and division operations, indicating positively changing the representation of two numerical labels introduces less offset of information to learn relations compared with negatively changing.
In addition, using either addition or multiplication may depend on the scale of the labels of the dataset. For example, if the labels are small (e.g., $<1$), it is expected that all multiplication pairs will become near 0, leading to ambiguity for model learning.
Moreover, division-based pre-training was the least consistent, often failing to converge during pre-training, as indicated by the ``-'' symbol.
This is because the divided changes are too significant to learn the relations.
These results highlight the adaptability of the arithmetic-aware pre-training method that is able to benefit different regression scenarios from various arithmetic operators.


%% file: src/conclusion.tex
This paper proposes APAR, a novel arithmetic-aware pre-training and adaptive-regularized fine-tuning framework for tabular regression tasks.
Distinct from existing works that ineffectively corrupt important features and transfer to regression labels due to the sparsity of the continuous space, our proposed pre-training task is able to take sample-wise interactions into account, allowing the capability of modeling from the aspects of continuous labels.
Meanwhile, our adaptive-regularized fine-tuning design dynamically adjusts appropriate data augmentation by conditioning it on the self-learned feature importance.
Experiments on 10 real-world datasets show that our APAR significantly outperforms state-of-the-art approaches by between 9.43\% to 20.37\%.
We believe that APAR serves as a generic framework for tabular regression applications due to the flexible design for pretrain-finetune frameworks, and multiple interesting directions could be further explored within the framework, such as automatically selecting appropriate arithmetic operations for effective pre-training, or extending APAR to classification tasks with Boolean operations (e.g., AND), etc.


%% file: src/appendix.tex
\section{The APAR Algorithm}
\subsection{Arithmetic-Aware Pre-Training}

The arithmetic-aware pre-training task aims to incorporate sample-wise information into each sample's representation in tabular regression tasks.
The process begins by randomly selecting two samples, which are processed through the feature tokenizer and feature encoder to generate their respective representations. The [CLS] token representations from both samples are then extracted, concatenated, and passed through an MLP to predict the outcome of the arithmetic operation.
The model is updated using stochastic gradient descent to minimize the MSE loss for this task.

\label{appendix:pretrain_algo}
\begin{algorithm}[tb]
	\caption{Arithmetic-Aware Pre-Training Phase} 
        \label{alg:pre-train}
        \textbf{Input:} Dataset $D=\{(x_i, y_i)\}_{i=1}^n$, mini-batch size $n_{mb}$, learning rate $\eta$ \\
        \textbf{Output:} Pre-trained parameters of \textit{Feature Tokenizer} $\theta_{\text{FT}}$ and \textit{Feature Encoder} $\theta_{\text{FE}}$
	\begin{algorithmic}[1]
            \STATE Initialize parameters of \textit{Feature Tokenizer} $\theta_{\text{FT}}$, \textit{Feature Encoder} $\theta_{\text{FE}}$ and a \textit{MLP} $\theta_{\text{MLP}}$
            \STATE Select an appropriate arithmetic operation $\vee \in \{+, -, \times, /\}$

            \REPEAT
                \STATE Set cumulative loss $\mathcal{L}^{\text{AP}} = 0$
                \FOR{$s = 1, \ldots, n_{mb}$}
                    \STATE Randomly draw two random samples with labels, ($x_i$, $y_i$) and ($x_j$, $y_j$), from $D$
                    \STATE Compute ground truth label for arithmetic operation:
                    \STATE \hspace{1cm} $y^{\text{AP}} = y_{i} \vee y_{j}$
                    \STATE Obtain feature embeddings for $x_i$ and $x_j$:
                    \STATE \hspace{1cm} $Z_i = \theta_{\text{FT}}(x_i)$
                    \STATE \hspace{1cm} $Z_j = \theta_{\text{FT}}(x_j)$
                    \STATE Generate representations for $x_i$ and $x_j$:
                    \STATE \hspace{1cm} $Z_{iL} = \theta_{\text{FE}}(\text{stack}[[\text{CLS}], Z_i])$
                    \STATE \hspace{1cm} $Z_{jL} = \theta_{\text{FE}}(\text{stack}[[\text{CLS}], Z_j])$
                    \STATE Predict arithmetic operation result:
                    \STATE \hspace{1cm} $\hat{y}^{\text{AP}} = \theta_{\text{MLP}}(\text{concat}[Z_{iL}^{[CLS]}, Z_{jL}^{[CLS]}])$
                    \STATE Calculate and accumulate loss:
                    \STATE \hspace{1cm} $\mathcal{L}^{\text{AP}} \leftarrow \mathcal{L}^{\text{AP}} + (y^{\text{AP}} - \hat{y}^{\text{AP}})^2$
                \ENDFOR

                \STATE Update \textit{Feature Tokenizer} $\theta_{\text{FT}}$:

                \STATE \hspace{1cm} $\theta_{\text{FT}} \leftarrow \theta_{\text{FT}} - \eta \nabla_{\theta_{\text{FT}}}(\frac{1}{n_{mb}}\mathcal{L}^{\text{AP}})$
                
                \STATE Update \textit{Feature Encoder} $\theta_{\text{FE}}$:

                \STATE \hspace{1cm} $\theta_{\text{FE}} \leftarrow \theta_{\text{FE}} - \eta \nabla_{\theta_{\text{FE}}}(\frac{1}{n_{mb}}\mathcal{L}^{\text{AP}})$

                \STATE Update \textit{MLP} $\theta_{\text{MLP}}$:
                
                \STATE \hspace{1cm} $\theta_{\text{MLP}} \leftarrow \theta_{\text{MLP}} - \eta \nabla_{\theta_{\text{MLP}}}(\frac{1}{n_{mb}}\mathcal{L}^{\text{AP}})$
                
            \STATE \UNTIL{convergence}
	\end{algorithmic} 
\end{algorithm}

\subsection{Relaxed Multivariate Bernoulli Distribution}
\label{appendix:fintune_bernoulli_algo}

In the fine-tuning phase of APAR, a relaxed multivariate Bernoulli distribution is used to generate a gate vector, which adaptively augments samples by considering feature importance and their correlated structure.
The process begins by drawing a sample from a standard normal distribution, followed by applying Cholesky decomposition to the correlation matrix derived from the training dataset.
This sample is then multiplied by the decomposed matrix to produce a Gaussian vector.
Finally, the Gaussian CDF and reparameterization trick are applied to each element of the vector to obtain the relaxed gate vector, which is multiplied with feature embeddings to augment the sample.

\begin{algorithm}[tb]
	\caption{Relaxed-MultiBern($\pi$; R) \protect\cite{wang2020relaxed}} 
        \label{alg:relaxed_multibern}
        \textbf{Input:} Selection probability $\pi$, correlation matrix $R$, temperature $\tau$ \\
        \textbf{Output:} Relaxed gate vector $\tilde{m}$
	\begin{algorithmic}[1]
            \STATE Draw a standard normal sample: $\epsilon \sim \mathcal{N}(0, I)$
            \STATE Compute $L =$ Cholesky-Decomposition$(R)$
            \STATE Generate a multivariate Gaussian vector $v = L\epsilon$

            \FOR{$i = 1, \ldots, k$}
                \STATE Apply element-wise Gaussian CDF: $u_i = \Phi(v_i)$
                \STATE Apply reparameterization trick:
                \STATE $\tilde{m}_i = \sigma\left(\frac{1}{\tau} (\log \pi_i - \log(1 - \pi_i) + \log u_i - \log(1 - u_i))\right)$
                \STATE where $\sigma(x) = \frac{1}{1+\exp(-x)}$.
            \ENDFOR
	\end{algorithmic} 
\end{algorithm}

\subsection{Adaptive-Regularized Fine-Tuning}
\label{appendix:fintune_algo}

The goal of the fine-tuning phase in APAR is to train the model to predict consistent labels from varying inputs, enabling model-guided data augmentation based on the learned importance of each feature, thereby improving performance on the downstream task.
Given an input sample, it is processed through the feature tokenizer and feature encoder to obtain its representation.
Simultaneously, the original feature embeddings are element-wise multiplied by the relaxed gate vector to create augmented feature embeddings.
These augmented embeddings are stacked with the [CLS] token and passed through the feature encoder to generate the augmented representation.
The [CLS] token representations from both the original and augmented samples are extracted and fed into an MLP to produce corresponding predictions.
The model is then updated using stochastic gradient descent to minimize the total MSE loss, which includes the target task, regularization, and feature sparsity.

\begin{algorithm}[tb]
	\caption{Adaptive-Regularized Fine-Tuning Phase} 
        \label{alg:fine-tune}
        \textbf{Input:} Dataset $D=\{(x_i, y_i)\}_{i=1}^n$, mini-batch size $n_{mb}$, learning rate $\eta$, loss weight for target task $\alpha$, regularized loss $\beta$ and sparsity loss $\gamma$, pre-trained parameters of \textit{Feature Tokenizer} $\theta_{\text{FT}}$ and \textit{Feature Encoder} $\theta_{\text{FE}}$ \\
        \textbf{Output:} Fine-tuned \textit{Feature Tokenizer} $\theta_{\text{FT}}$,  \textit{Feature Encoder} $\theta_{\text{FE}}$ and \textit{MLP} $\theta_{\text{MLP}}$
	\begin{algorithmic}[1]
            
            \STATE Initialize parameters of \textit{MLP} $\theta_{\text{MLP}}$
            \STATE Initialize selection probability $\pi$
            \STATE Calculate correlation matrix $R$ from $D$ 

            \REPEAT

                \STATE Set cumulative loss $\mathcal{L}^{\text{AR}} = 0$
                \FOR{$s = 1, \ldots, n_{mb}$}
                    \STATE Randomly draw a random sample with label, ($x_i$, $y_i$), from $D$
                    
                    \STATE Obtain feature embeddings for $x_i$:
                    \STATE \hspace{1cm} $Z_i = \theta_{\text{FT}}(x_i)$

                    \STATE Generate relaxed gate vector:
                    \STATE \hspace{1cm} $\tilde{m} \sim Relaxed$-$MultiBern(\pi; R)$

                    \STATE Obtain augmented feature embeddings:
                    \STATE \hspace{1cm} $\tilde{Z}_{i} = Z_{i} \odot \tilde{m}$

                    \STATE Generate representations for original and augmented sample:
                    \STATE \hspace{1cm} $Z_{iL} = \theta_{\text{FE}}(\text{stack}[[\text{CLS}], Z_i])$
                    \STATE \hspace{1cm} $\tilde{Z}_{iL} = \theta_{\text{FE}}(\text{stack}[[\text{CLS}], \tilde{Z}_{i}])$

                    \STATE Generate the predictions for target task:
                    \STATE \hspace{1cm} $\hat{y}_{i}^{\text{AR}} = \theta_{\text{MLP}}(Z_{iL}^{[\text{CLS}]})$
                    \STATE \hspace{1cm} $\tilde{y}_{i}^{\text{AR}} = \theta_{\text{MLP}}(\tilde{Z}_{iL}^{[\text{CLS}]})$

                    \STATE Calculate and accumulate loss:
                    \STATE \hspace{1cm} $\mathcal{L}^{\text{AR}} \leftarrow \mathcal{L}^{\text{AR}} + \alpha \cdot (y_i - \hat{y}_{i}^{\text{AR}})^2 + \beta \cdot (y_i - \tilde{y}_{i}^{\text{AR}})^2 + \gamma \cdot \sum_{i}^{k} \pi_{i}$
                \ENDFOR

                \STATE Update \textit{Feature Tokenizer} $\theta_{\text{FT}}$:

                \STATE \hspace{1cm} $\theta_{\text{FT}} \leftarrow \theta_{\text{FT}} - \eta \nabla_{\theta_{\text{FT}}}(\frac{1}{n_{mb}}\mathcal{L}^{\text{AR}})$
                
                \STATE Update \textit{Feature Encoder} $\theta_{\text{FE}}$:

                \STATE \hspace{1cm} $\theta_{\text{FE}} \leftarrow \theta_{\text{FE}} - \eta \nabla_{\theta_{\text{FE}}}(\frac{1}{n_{mb}}\mathcal{L}^{\text{AR}})$

                \STATE Update \textit{MLP} $\theta_{\text{MLP}}$:
                
                \STATE \hspace{1cm} $\theta_{\text{MLP}} \leftarrow \theta_{\text{MLP}} - \eta \nabla_{\theta_{\text{MLP}}}(\frac{1}{n_{mb}}\mathcal{L}^{\text{AR}})$

            \STATE \UNTIL{convergence}
	\end{algorithmic} 
\end{algorithm}

\section{Details and Setups of Baselines}
\label{appendix:basline_setup}


We used the default settings of XGBoost \cite{chen2016xgboost}, LightGBM \cite{ke2017lightgbm}, and CatBoost \cite{prokhorenkova2018catboost}, which represent three established GBDT-based models in the tabular data field, based on the official implementation\footnote{\url{https://github.com/dmlc/xgboost/tree/stable}}\footnote{\url{https://github.com/microsoft/LightGBM}}\footnote{\url{https://github.com/catboost/catboost}}, implementing early stopping rounds of 5 and Optuna for tuning. The search spaces are: iterations \{100 to 500\}, depth \{6 to 8\}, and learning rate \{5$e$-5 to 5$e$-1\}. In the tree-based models, we found that tree depths from 4 to 12 and iterations from 100 to 1,000 showed performance gains plateauing beyond 500 iterations. Depths from 4 to 8 improved performance, but depths beyond 8 led to earlier termination and slight performance decline. Hence, our search space for tree depths is 6 to 8 and iterations 100 to 500. Our analysis showed no underfitting or overfitting is due to adequate tree depth, iterations, and subsample and early stopping techniques. The MLP, which serves as a fundamental baseline with a straightforward deep learning structure for tabular data, consists of 8 blocks, each with a linear layer and ReLU activation, a hidden dimension of 512, and an initial learning rate of 5$e$-4, mirroring APAR's scheduler.

For AutoInt \cite{song2019autoint}, which utilizes embeddings and self-attention mechanisms for tabular data feature transformation, we adjust the hidden dimension \{16 to 256\}, attention blocks \{3 to 5\}, and MHSA heads \{2 to 6\} for better results based on the open source package\footnote{\url{https://github.com/manujosephv/pytorch_tabular}}. FT-Transformer \cite{gorishniy2021revisiting}, which adapts the Transformer architecture incorporating a feature tokenizer for tabular data follows the official implementation \footnote{\url{https://github.com/yandex-research/rtdl}} but tunes the number of Transformer layers \{1 to 6\}. TabNet \cite{arik2021tabnet}, which employs a recurrent structure that combines dynamic feature reweighing with traditional feed-forward components, uses the official implementation\footnote{\url{https://github.com/google-research/google-research/tree/master/tabnet}}, fine-tuning decision steps \{1 to 128\} and virtual batch size \{128 to 512\}. VIME \cite{yoon2020vime}, which introduces methods for both self-supervised and semi-supervised learning in tabular contexts, maintains a consistent configuration of the MLP architecture across datasets. The pre-training phase tunes the feature masking probability \{1$e$-1 to 1$e$0\} and loss weight \{1 to 2\}, while the fine-tuning phase adjusts the feature masking probability \{1$e$-2 to 1$e$0\} and the number of augmented samples \{3 to 5\}.

For APAR, we use default hyperparameters from previous research \cite{gorishniy2021revisiting}: 3 Transformer layers, a feature embedding size of 192, and 8 heads in the Multi-Head Self-Attention (MHSA). Dropout rates are 0.2 for attention, 0.1 for the Feed-Forward Network (FFN), and 0.0 for residual components. We initialized the model using the Kaiming method. We predefined the configuration set \{0.010, 0.025, 0.050, 0.075, 0.1, 0.2, 0.3, 0.4, 0.5\} for the weights of the regularized loss $\mathcal{L}_{\text{reg}}$ and sparsity loss $\mathcal{L}_{\text{sparsity}}$ ($\beta$ and $\gamma$). Initial learning rates were 1$e$-3 for pre-training and 5$e$-4 for fine-tuning, using the StepLR scheduler with a decay rate of 0.98.

\section{Details of Datasets}
\label{appendix:dataset}

The source and tasks of the ten datasets used in our experiment are as follows:

\begin{itemize}
\item Taiwan Housing (BD, AM, HS) \cite{tw_real_estate_platform}: This dataset consists of three building types (building, apartment, and house) and is used to predict property prices \footnote{\url{https://lvr.land.moi.gov.tw/}}.
\item Gas Emission (GS) \cite{misc_gas_turbine_co_and_nox_emission_data_set_551}: This dataset is used to predict flue gas emissions in Turkey \footnote{\url{https://archive.ics.uci.edu/dataset/551/gas+turbine+co+and+nox+emission+data+set}}.
\item Election Results (ER) \cite{misc_real-time_election_results:_portugal_2019_513}: This dataset is aimed at predicting the results of the 2019 Portuguese Parliamentary Election \footnote{\url{https://archive.ics.uci.edu/dataset/513/real+time+election+results+portugal+2019}}.
\item Beijing PM2.5 (PM) \cite{misc_beijing_pm2.5_381}: This dataset is used to predict hourly PM2.5 levels in Beijing \footnote{\url{https://archive.ics.uci.edu/dataset/381/beijing+pm2+5+data}}.
\item Bike Sharing (BS) \cite{misc_bike_sharing_dataset_275}: This dataset aims to predict the hourly and daily count of rental bikes between 2011 and 2012 in the Capital Bikeshare system \footnote{\url{https://archive.ics.uci.edu/dataset/275/bike+sharing+dataset}}.
\item Year (YE) \cite{misc_yearpredictionmsd_203}: This dataset is used to predict the release year of a song based on audio features \footnote{\url{https://archive.ics.uci.edu/dataset/203/yearpredictionmsd}}.
\item Kernel Performance (KP) \cite{misc_sgemm_gpu_kernel_performance_440}: This dataset is used to predict the running times for multiplying two 2048 x 2048 matrices using a GPU OpenCL SGEMM kernel \footnote{\url{https://archive.ics.uci.edu/dataset/440/sgemm+gpu+kernel+performance}}.
\item Flight Price (FP) \cite{flight_price_kaggle}: This dataset is used to predict flight prices from the "Ease My Trip" website \footnote{\url{https://www.kaggle.com/datasets/shubhambathwal/flight-price-prediction}}.
\end{itemize}

\section{Additional Experiments}
\label{appendix:additional_experiment}

To further validate the effectiveness of our proposed APAR method, we conducted experiments with three additional baselines: TabTransformer \cite{huang2020tabtransformer}, a classical and widely-recognized approach for tabular data; TabR \cite{gorishniy2024tabr}, a recently proposed state-of-the-art method for tabular regression; and Mambular \cite{thielmann2024mambular}, another competitive recent baseline. We evaluated these methods across three datasets: ER, BS, and PM. The results in Table~\ref{tab:additional_experiment} demonstrate the effectiveness of APAR in achieving superior performance compared to the baselines.

\begin{table}[h]
\centering
\caption{Quantitative results of additional baselines and our proposed APAR. For each dataset, the best result in each column is in boldface.}
\label{tab:additional_experiment}
\begin{tabular}{lccc}
\toprule
\textbf{}       & \textbf{ER} & \textbf{BS} & \textbf{PM} \\ \midrule
TabTransformer & 0.094       & 0.203       & 0.945       \\
TabR & 0.073       & 0.067       & 0.527       \\
Mambular & 0.063       & 0.054       & 0.548       \\
\textbf{APAR (Ours)}    & \textbf{0.034} & \textbf{0.014} & \textbf{0.524} \\
\bottomrule
\end{tabular}
\end{table}